\documentclass[pdflatex,sn-mathphys]{sn-jnl}% Math and Physical Sciences Reference Style
\jyear{2021}%

\theoremstyle{thmstyleone}%
\theoremstyle{thmstyletwo}%

\theoremstyle{thmstylethree}%

\begin{document}

\title[Article Title]{Detecting Disengagement in Virtual Learning as an Anomaly using Temporal Convolutional Network Autoencoder}


\author*[1]{\fnm{Ali} \sur{Abedi}}\email{ali.abedi@uhn.ca}

\author[1]{\fnm{Shehroz} \sur{S. Khan}}\email{shehroz.khan@uhn.ca}

\affil[1]{\orgdiv{KITE}, \orgname{University Health Network, Canada}}


\abstract{Student engagement is an important factor in meeting the goals of virtual learning programs. Automatic measurement of student engagement provides helpful information for instructors to meet learning program objectives and individualize program delivery. Many existing approaches solve video-based engagement measurement using the traditional frameworks of binary classification (classifying video snippets into engaged or disengaged classes), multi-class classification (classifying video snippets into multiple classes corresponding to different levels of engagement), or regression (estimating a continuous value corresponding to the level of engagement). However, we observe that while the engagement behaviour is mostly well-defined (e.g., focused, not distracted), disengagement can be expressed in various ways. In addition, in some cases, the data for disengaged classes may not be sufficient to train generalizable binary or multi-class classifiers. To handle this situation, in this paper, for the first time, we formulate detecting disengagement in virtual learning as an anomaly detection problem. We design various autoencoders, including temporal convolutional network autoencoder, long-short-term memory autoencoder, and feedforward autoencoder using different behavioral and affect features for video-based student disengagement detection. The result of our experiments on two publicly available student engagement datasets, DAiSEE and EmotiW, shows the superiority of the proposed approach for disengagement detection as an anomaly compared to binary classifiers for classifying videos into engaged versus disengaged classes (with an average improvement of $9$\% on the area under the curve of the receiver operating characteristic curve and $22$\% on the area under the curve of the precision-recall curve).}




\maketitle

\section{Introduction}\label{sec1}
\label{sec:introduction}
With the widespread accessibility and adoption of internet services across major urban centers and universities, virtual learning programs are becoming more ubiquitous and mainstream \cite{dhawan2020online}. Virtual learning programs offer many advantages compared to traditional in-person learning programs in terms of being more accessible, economical, and personalizable \cite{dumford2018online}. However, virtual learning programs also bring other types of challenges. For instance, in a virtual learning setting, students and tutor are behind a “virtual wall”, and it becomes very difficult for the tutor to assess the students’ engagement in the class being taught. This problem is further intensified if the group of students is large \cite{dumford2018online,sumer2021multimodal}. Therefore, from a tutor’s perspective, it is crucial to automatically measure student engagement to provide them with real-time feedback and take necessary actions to engage the students to maximize their learning objectives.

According to Sinatra et al. \cite{sinatra2015challenges}, in engagement measurement, the focus is on the behavioral, affective, and cognitive states of the student in the moment of interaction with a particular contextual environment. Engagement is not stable over time and is best captured with physiological and psychological measures at fine-grained time scales, from seconds to minutes \cite{d2017advanced}. Behavioral engagement involves general on-task behavior and paying attention at the surface level. The indicators of behavioral engagement, in the moment of interaction, include eye contact, blink rate, and head pose \cite{d2017advanced,woolf2009affect}. Affective engagement is defined as the affective and emotional reactions of the student to the content. Its indicators are positive versus negative and activating versus deactivating emotions \cite{sinatra2015challenges,d2017advanced,woolf2009affect}. Cognitive engagement pertains to the psychological investment and effort allocation of the student to deeply understand the learning materials \cite{sinatra2015challenges}. To measure cognitive engagement, information such as student’s speech could be processed to recognize the level of student’s comprehension of the context \cite{sinatra2015challenges}. Contrary to behavioral and affective engagements, measuring cognitive engagement requires knowledge about context materials. Measuring student’s engagement in a specific context depends on the knowledge about the student and the context. From a data analysis perspective, it depends on the data modalities available to analyze. In this paper, the focus is on automatic video-based disengagement detection. The only data modality is video, without audio, and with no knowledge about the context.

Majority of the recent works on student engagement measurement are based on the video data of students acquired by cameras and using deep-learning, machine-learning, and computer-vision techniques \cite{dewan2019engagement,doherty2018engagement,abedi2021improving}. Previous works on student engagement measurement tried to solve different machine-learning problems, including

\begin{itemize}
    \item Binary classification problem \cite{aung2018harnessing,chen2019faceengage,booth2017toward,gupta2016daisee, mehta2022three}, classifying student’s video into engaged or disengaged classes,
    \item Multi-class classification problem \cite{abedi2021affect,booth2017toward,gupta2016daisee,liao2021deep,abedi2021improving,huang2019fine,ma2021automatic,dresvyanskiy2021deep, mehta2022three}, classifying student’s video into multiple classes corresponding to different levels of engagement, or
    \item Regression \cite{abedi2021affect,booth2017toward,liao2021deep,thomas2018predicting,kaur2018prediction,copur2022engagement}, estimating a continuous value corresponding to the level of engagement of the student in the video.
\end{itemize}

\begin{figure*}
    \centering
    \includegraphics[width=\linewidth]{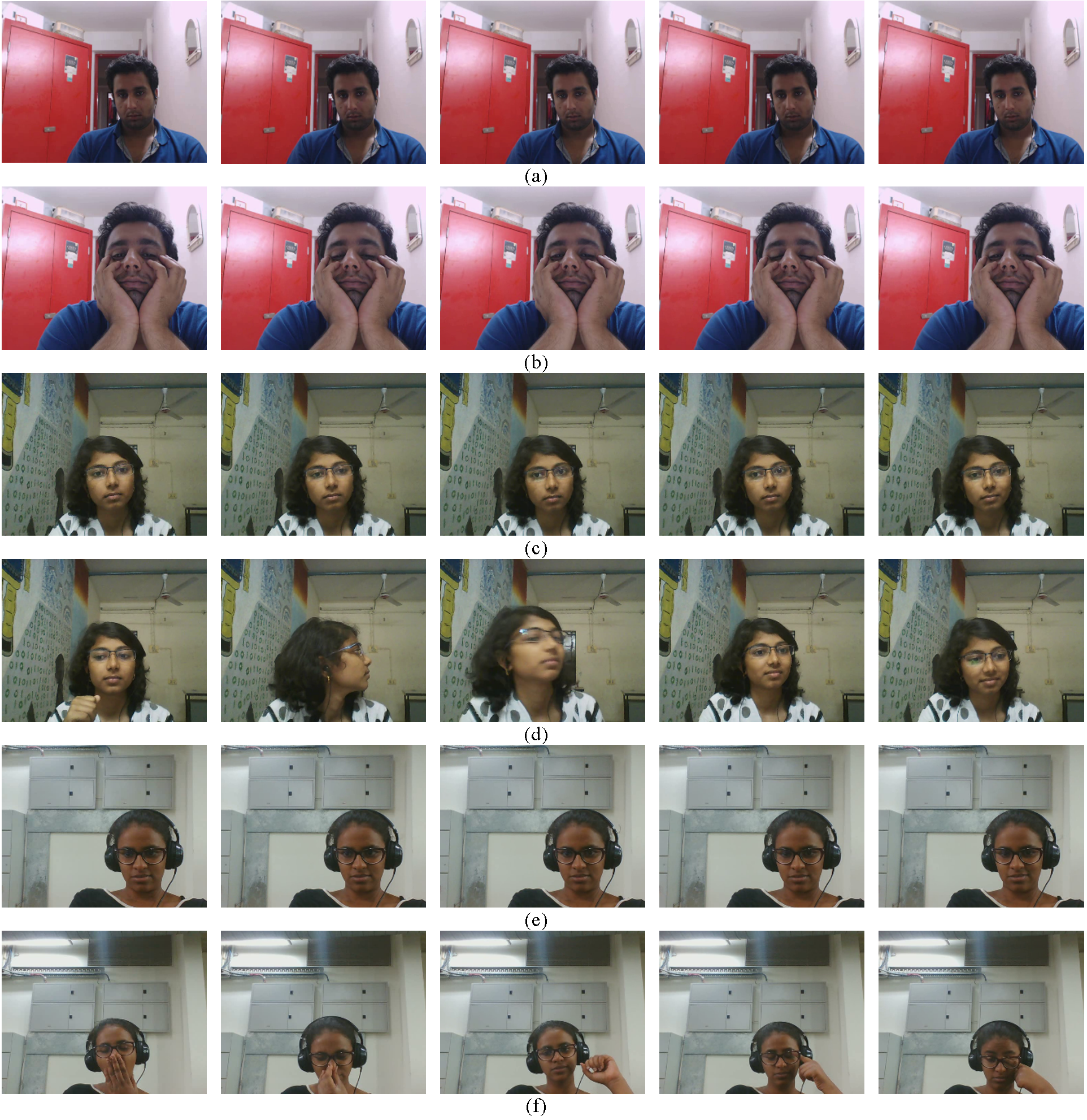}
\caption{5 (out of 300) frames of six video samples of three students in the DAiSEE dataset \cite{gupta2016daisee}. While the students in (a), (c), and (e) are engaged, the students in (b), (d), and (f) are disengaged. As can be seen in these exemplary frames, while the students in the engaged videos show similar behavioural and affective states (attentively looking at the camera with relatively positive affect states), the behavioural and affective indicators of disengagement are different in the disengaged students, face-palming and sleepy affect state in (b), looking somewhere other than camera in (d), yawning, eye rubbing, and eye closure in (f). The proposed algorithm for disengagement detection, temporal convolutional autoencoder with behavioural and affect features after applying a threshold on its output reconstruction error (described in Section \ref{sec:methodology}), successfully detects disengagement in (b), (d), and (f) (true positive). It correctly classifies the students in (a), (c), and (e) into engaged (true negative).}
\label{fig:exemplary}
\end{figure*}

The state of high and medium engagement is mostly well understood and well defined in terms of behavioral and affective states; if a student looking at the camera with attention and having a focused affect state is considered engaged. However, disengagement could be expressed in diverse ways, including various combinations of behavioral and affective states, such as off-task behavioral states; not looking at the camera \cite{sinatra2015challenges,d2017advanced,woolf2009affect,aslan2017human}, high blink rate \cite{ranti2020blink}, and face-palming, or negative and deactivating affect states; annoyed (high negative valence and high positive arousal), and bored (high negative valence and high negative arousal) emotional states \cite{sinatra2015challenges,d2017advanced,woolf2009affect,aslan2017human}, see Figure \ref{fig:exemplary} for some examples. Collecting large amounts of data corresponding to these diverse disengagement states is challenging. As a consequence, in many engagement measurement datasets, data corresponding to different types of disengagement states could be sparse \cite{khan2022inconsistencies}, and building supervised models on a diverse class of "disengagement" can be very difficult \cite{dresvyanskiy2021deep,liao2021deep,abedi2021improving}. In this paper, for the first time, we formulate student disengagement detection as an anomaly detection problem. This formulation does not imply that disengagement is a rare event, rather it is meant to detect the diversity and lack of consistency in expressing disengagement in virtual learning. Using an anomaly detection framework, those behavioral and affect states that significantly deviate from a well-defined engaged state can be identified as the disengaged state. The anomaly detection frameworks to detect (not necessarily rare) deviant behaviors from normal behaviors have been successfully exploited in other domains, such as driver anomalous behavior detection \cite{kopuklu2021driver,khan2021modified}.

We also observed that in many existing student engagement datasets, the distribution of engaged to disengaged samples is highly imbalanced; in many cases, the percentage of disengaged samples is extremely low \cite{dresvyanskiy2021deep,khan2022inconsistencies}. The interpretation of this imbalanced data distribution in educational psychology is that the students are more engaged in the courses that are relevant to their major \cite{barlow2020development,canziani2021student}. In this severely skewed disengagement class scenario, supervised approaches using weighted loss functions could be helpful \cite{abedi2021improving}. However, according to the poor disengagement class accuracies reported in previous works \cite{liao2021deep,mehta2022three,abedi2021improving}, the diversity of disengaged states is challenging to be modeled, and the anomaly detection approach is a viable approach to detect these states.

Based on the above discussion of anomaly detection framework to detect disengaged behaviors, we design various autoencoder (AE) neural networks to detect student disengagement in videos. Different behavioral and affect features, as indicators of behavioral and affective components of engagement \cite{sinatra2015challenges,d2017advanced,woolf2009affect,aslan2017human} are extracted from videos. Various Temporal Convolutional Network (TCN) \cite{thill2021temporal}, Long-Short-Term Memory (LSTM), and feedforward AEs using the extracted features are designed for video-based student disengagement detection. The AEs are trained on only engaged video snippets that are abundantly available. It is expected that an AE will be able to model the "engaged" concept through low reconstruction error. An unseen "disengaged" video snippet would result in a larger reconstruction error indicating deviance from the "engaged" class. Thus the reconstruction error can be thresholded to detect disengagement. The developed temporal and convolutional network AEs analyze the sequences of features extracted from consecutive video frames. Therefore, in addition to being able to detect anomalous behavioral and affective states (static indicators of disengagement), they are able to detect anomalous temporal changes in the behavioral and affective states (dynamic indicators of disengagement) \cite{d2012dynamics}. The experimental result on two publicly available student engagement datasets shows that the anomaly detection approach outperforms the equivalent binary classification formulation. Our primary contributions are as follows:
\begin{itemize}
    \item This is the first work formulating disengagement detection as an anomaly detection problem and developing AEs for this task \cite{dewan2019engagement,doherty2018engagement,khan2022inconsistencies}.
    \item Temporal Convolutional Network AE (TCN-AE) has recently been introduced for unsupervised anomaly detection in time series \cite{thill2021temporal}. For the first time in the field of affective computing, TCN-AE is used for disengagement detection.
    \item Extensive experiments are conducted on the only two publicly available student engagement datasets \cite{gupta2016daisee,kaur2018prediction}, and the proposed anomaly detection formulation is compared with various feature-based and end-to-end binary classification methods.
\end{itemize}
This paper is structured as follows. Section \ref{sec:literature_review} introduces related works on student engagement measurement, focusing on the machine-learning problem they solve. In Section \ref{sec:methodology}, the proposed approach for student disengagement detection is presented. Section \ref{sec:experiments} describes experimental settings and results on the proposed methodology. In the end, Section \ref{sec:conclusion} presents our conclusions and directions for future works.

\section{Literature Review}
\label{sec:literature_review}
Over the recent years, extensive research efforts have been devoted to automating student engagement measurement \cite{dewan2019engagement,doherty2018engagement,abedi2021affect}. This section briefly discusses previous works on computer-vision-based engagement measurement, focusing on their machine-learning problem. For more information on features, machine-learning algorithms, deep neural network architectures, and datasets of the previous methods, refer to \cite{dewan2019engagement,doherty2018engagement,abedi2021affect,khan2022inconsistencies}.

\begin{table*}[]
% \tiny
\caption{Previous video-based engagement measurement approaches compared to the proposed approach in this paper, the machine-learning problem they solve (Binary Classification: BC, Multi-class Classification: MC, or Regression: R), their features, and their machine-learning/deep-learning models.}
\label{tab:literature}
\centering
\begin{tabular}{p{.2\linewidth}p{.1\linewidth}p{.4\linewidth}p{.2\linewidth}}
\hline
Authors & Problem & Features & Model\\
\hline
Gupta et el., 2016 \cite{gupta2016daisee} & MC, BC	& end-to-end	& C3D, LSTM
\\
\hline
Bosch et al. 2016 \cite{bosch2016detecting} & BC & AU, LBP-TOP, Gabor & SVM
\\
\hline
Booth et al., 2017 \cite{booth2017toward} & BC, MC, R	& facial landmark, AU, optical flow, head pose	& SVM, KNN, RF
\\
\hline
Aung et al., 2018 \cite{aung2018harnessing} & MC, R & box filter, Gabor filter, facial Action Units (AU) & GentleBoost, SVM, LR
\\
\hline
Niu et al., 2018 \cite{niu2018automatic} & R & gaze, head pose, AU & GRU
\\
\hline
Thomas et al., 2018 \cite{thomas2018predicting} & R &	gaze, head pose, AU	& TCN
\\
\hline
Huang et al., 2019 \cite{huang2019fine} & MC	& gaze, eye location, head pose, AU	& LSTM
\\
\hline
Kaur et al., 2018 \cite{kaur2018prediction} & R	& LBP-TOP	& MLP
\\
\hline
Chen et al., 2019 \cite{chen2019faceengage} & BC	& gaze, blink rate, head pose, facial embedding	& SVM, KNN, RF, RNN
\\
\hline
Wu et al., 2020 \cite{wu2020advanced} & R	& gaze, head   pose, body pose, facial embedding	& LSTM, GRU
\\
\hline
Liao et al., 2021 \cite{liao2021deep} & MC, R	& gaze, eye location, head pose	& LSTM
\\
\hline
Ma et al., 2021 \cite{ma2021automatic} & MC & gaze direction, head pose, AU, C3D	& neural Turing machine
\\
\hline
Abedi et al., 2021 \cite{abedi2021affect} & MC, R	& affect, facial embedding, blink rate, gaze head pose	& LSTM, TCN, MLP, SVM, RF
\\
\hline
Dresvyanskiy et al., 2021 \cite{dresvyanskiy2021deep} & MC	& AU, facial embedding	& LSTM
\\
\hline
Abedi et al., 2021 \cite{abedi2021improving} & MC	& end-to-end	& ResNet + TCN
\\
\hline
Copur et al. 2022 \cite{copur2022engagement} & R & eyae gaze, head pose, AU & LSTM
\\
\hline
\textbf{proposed method}	& \textbf{anomaly detection}	& \textbf{valence, arousal, blink rate, gaze, head pose}	& \textbf{TCN-AE, LSTM-AE}
\\
\hline
\end{tabular}
\end{table*}

Table \ref{tab:literature} shows previous related works on video-based engagement measurement and the machine-learning problem they solve. The majority of previous works on engagement measurement solve three types of problems: binary classification, multi-class classification, and regression.

According to the datasets used in the previous approaches, the authors developed different machine-learning algorithms or deep neural-network architectures to measure engagement. Two publicly available video-based engagement measurement datasets are (i) DAiSEE (Dataset for Affective States in E-Environments) \cite{gupta2016daisee} in which engagement measurement is defined as a four-class classification problem, and (ii) EmotiW (Emotion Recognition in the Wild) \cite{kaur2018prediction} in which engagement measurement is defined as a regression problem. The approaches that use DAiSEE mostly consider it as a four, three, or two-class classification problem \cite{abedi2021affect,gupta2016daisee,liao2021deep,abedi2021improving,huang2019fine,ma2021automatic,dresvyanskiy2021deep}, whereas the approaches that use EmotiW apply various regression techniques to model the level of engagement \cite{abedi2021affect,liao2021deep,thomas2018predicting,kaur2018prediction,niu2018automatic,wu2020advanced}. Many authors altered the original machine-learning problem in these two datasets and designed models based on the new problem, such as converting the original four-class classification problem in DAiSEE into a regression problem and developing regression deep-learning models \cite{liao2021deep}. Other previous methods defined their models based on the problems in their non-public datasets. For instance, Aung et al., \cite{aung2018harnessing} collected a video dataset for student engagement measurement and designed different machine-learning models to solve a four-class classification problem and a regression problem. In the end-to-end approaches in Table \ref{tab:literature} \cite{gupta2016daisee,abedi2021improving}, there is no feature extraction step, and the deep-learning models output the engagement level of raw video frames as inputs. Abedi and Khan \cite{abedi2021affect} proposed to use affect features (continuous values of valence and arousal) along with behavioral features for engagement level classification and regression. In this paper, using these handcrafted features, we propose to use AE neural networks for student disengagement detection in videos.

\begin{figure*}
\label{fig:methodology}
    \centering
    \includegraphics[width=\linewidth]{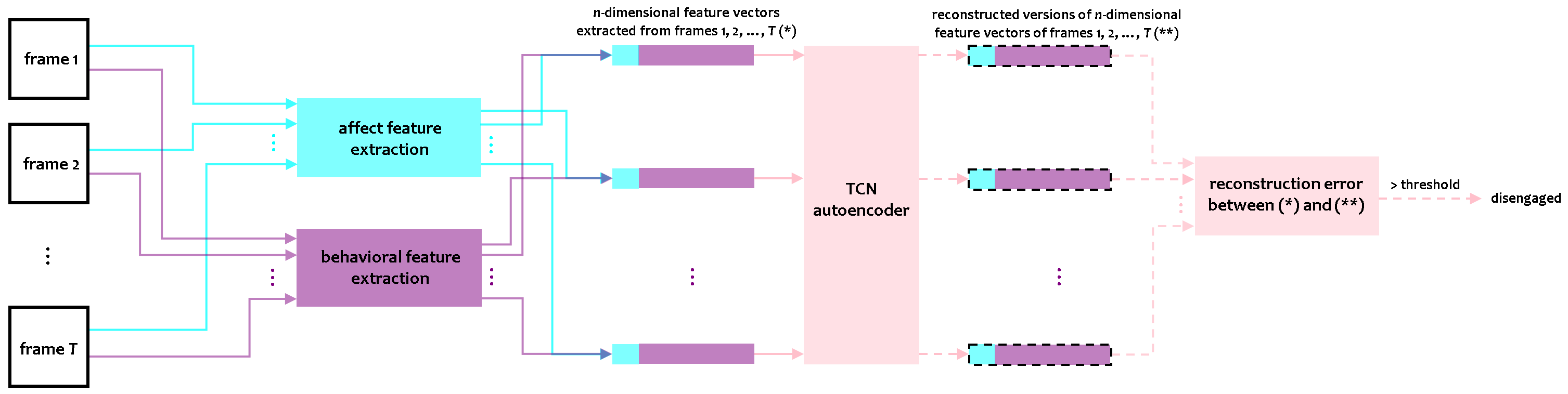}
\caption{Block diagram of the proposed method for student disengagement detection using TCN-AE. $n$-dimensional affect and behavioral feature vectors are extracted from $T$ consecutive input video frames and fed to TCN-AE. In the training phase, the TCN-AE is trained on the videos of engaged students to reconstruct their feature vectors (the dashed rectangles) by minimizing the reconstruction error. In the test phase, the disengagement of the student in the input video is detected by applying a threshold on the output reconstruction error of the trained TCN-AE, see Section \ref{sec:methodology}. In another setting, explained in Section \ref{sec:methodology}, the input to the method can be $T$ consecutive video segments instead of $T$ consecutive video frames.}
\end{figure*}

\section{Methodology}
\label{sec:methodology}
Figure \ref{fig:methodology} depicts the conceptual block diagram of the proposed method for student disengagement detection using AEs. The input is a video clip of a student in a virtual learning session sitting in front of the camera of a laptop or PC, see Figure \ref{fig:exemplary}. Various behavioral and affective features are extracted from the consecutive frames of the input video to construct a feature vector as the input to a neural network. The neural network model is an AE that has been trained on the feature vectors of the videos of engaged students (normal samples). The AE outputs the error of reconstructing the input video. This reconstruction error shows how much the student in the input video clip is disengaged, i.e., how much the behavioral and affective states of the student are deviant from the states of an engaged student (a normal sample).

\subsection{Feature Extraction}
\label{sec:feature_extraction}
Various affect and behavioral features, corresponding to the affective and behavioral states of the student in the video, are extracted from the consecutive video frames.

\textbf{Affect features:} Altuwairqi et al. \cite{altuwairqi2021new} linked engagement to affective states. They defined different levels of engagement according to different values of valence and arousal in the circumplex model of affect \cite{altuwairqi2021new}. Abedi et al. \cite{abedi2021affect} showed that sequences of continuous values of valence and arousal extracted from consecutive video frames are significant indicators of affective engagement. In the circumplex model of affect \cite{altuwairqi2021new}, different combinations of values of valence and arousal correspond to different affect states. Positive values of valence and arousal with low fluctuations over time correspond to affective engagement, and any deviation from these values indicates affective disengagement. The pretrained EmoFAN \cite{toisoul2021estimation} on the AffectNet dataset \citep{mollahosseini2017affectnet}, a deep neural network to analyze facial affects in face images, is used to extract valence and arousal from consecutive video frames (refer to \cite{abedi2021affect} for more detail).

\textbf{Behavioral features:} Woolf et al. \cite{woolf2009affect} defined different characteristics of behavioral engagement as on-task behavior against off-task behavior corresponding to behavioral disengagement. In many cases, a student engaged in a virtual learning material focusedly looks at the computer screen, i.e., the student’s head pose and eye gaze direction are perpendicular to the computer screen with low fluctuations. There are low fluctuations in yaw, pitch, and roll of the student’s head while engaged with the learning material. Accordingly, inspired by previous research \cite{abedi2021affect,liao2021deep,thomas2018predicting,kaur2018prediction,niu2018automatic}, eye location, head pose, and eye gaze direction in consecutive video frames are considered as one set of behavioral features. Ranti et al. \cite{ranti2020blink} demonstrated that eye blinks are withdrawn at precise moments in time so as to minimize the loss of visual information that occurs during a blink. High and irregular blink rate indicates disengagement \cite{niu2018automatic,ranti2020blink}. The intensity of facial action unit AU45, indicating how closed the eyes are \cite{abedi2021affect}, will be used as another behavioral feature.

According to the above explanations, the following 11 features are extracted from consecutive video frames, 2-element affect features: valence and arousal (2 features), and 9-element behavioral features: eye-closure intensity (1 feature), x and y components of eye gaze direction w.r.t. camera (2 features), x, y, and z components of head location w.r.t. camera (3 features); pitch, yaw, and roll of the head (3 features).

\begin{figure*}
    \centering
    \includegraphics[scale=.2]{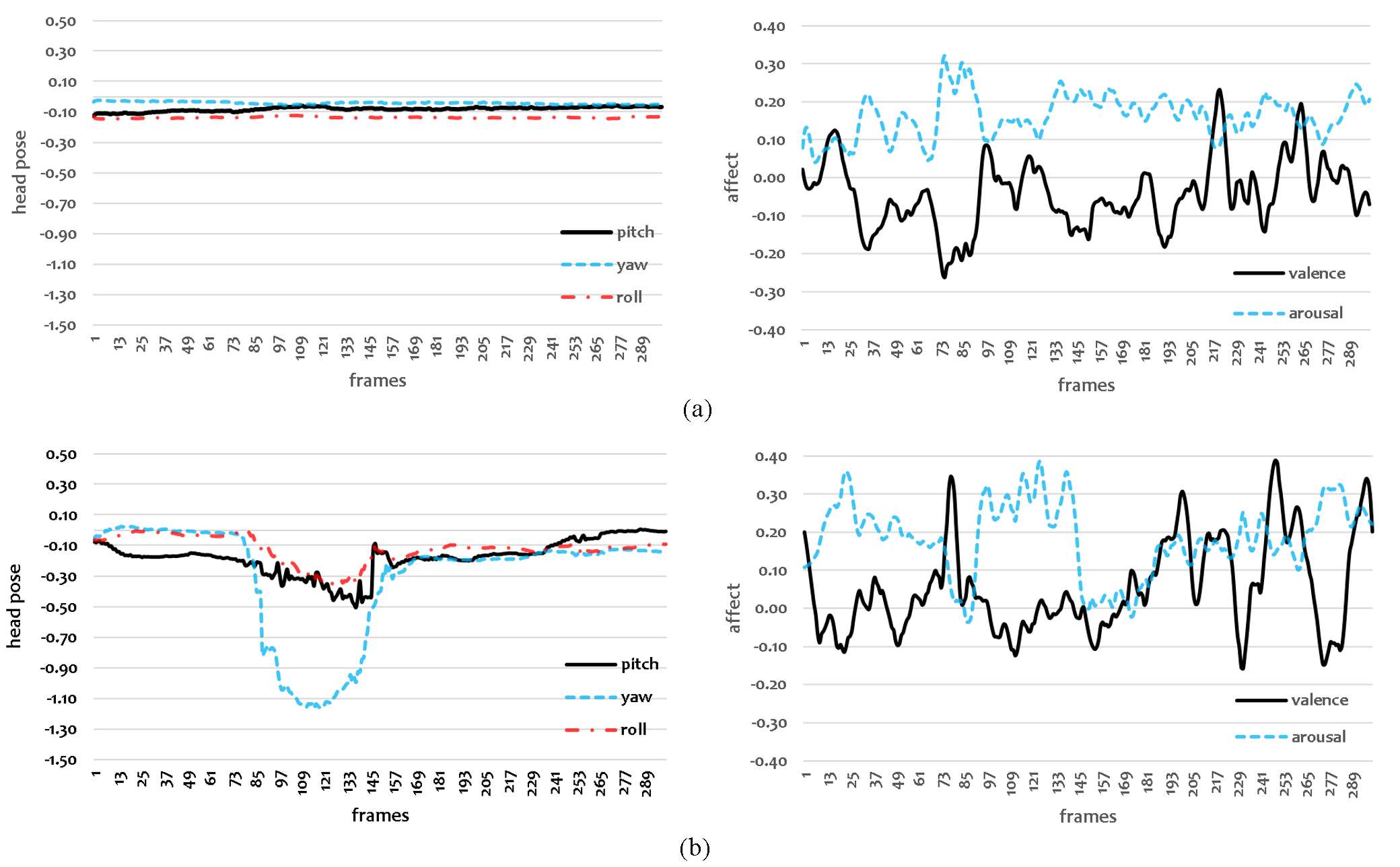}
\caption{The pitch, yaw, and roll of the head (as examples of the behavioral features) and valence and arousal (as affect features) extracted from the 300 consecutive frames of (a) the engaged sample in Figure \ref{fig:exemplary} (c) and (b) the disengaged sample in Figure \ref{fig:exemplary} (d). The values of features over consecutive frames are more fluctuating and anomalous in the disengaged sample helping AEs to detect disengagement.}
\label{fig:features}
\end{figure*}

To illustrate how the extracted behavioral and affect features are capable of differentiating between engaged and disengaged samples, some of the extracted features from the consecutive 300 frames of the videos in Figures \ref{fig:exemplary} (c) and (d) are drawn in Figure \ref{fig:features}. The difference between features extracted from the video in Figure \ref{fig:exemplary} (c), an engaged sample, and Figure \ref{fig:exemplary} (d), a disengaged sample, is observable in Figure \ref{fig:features}. These differentiating features and anomalies in features of the disengaged sample over time help AEs detect disengagement.

In addition to using the above 11-element \textit{frame-level features}, being extracted from single frames of video, 37-element \textit{segment-level features} are also extracted from video segments. Each video segment consists of multiple consecutive video frames. The mean and standard deviation of valence and arousal values over consecutive video frames (4 features), the blink rate, derived by counting the number of peaks above a certain threshold divided by the number of frames in the AU45 intensity time series extracted from the input video (1 feature), the mean and standard deviation of the velocity and acceleration of x and y components of eye gaze direction (8 features), the mean and standard deviation of the velocity and acceleration of x, y,  and z components of eye gaze direction (12 features), and the mean and standard deviation of the velocity and acceleration of head’s pitch, yaw, and roll (12 features). As will be described in Section \ref{sec:experiments}, the frame-level, and segment-level features are used to extract features from short, and long videos, respectively.

\subsection{Video-based Disengagement Detection Using Temporal Convolutional Network Autoencoder}
\label{sec:anomaly_detection}
\begin{figure*}
    \centering
    \includegraphics[scale=.25]{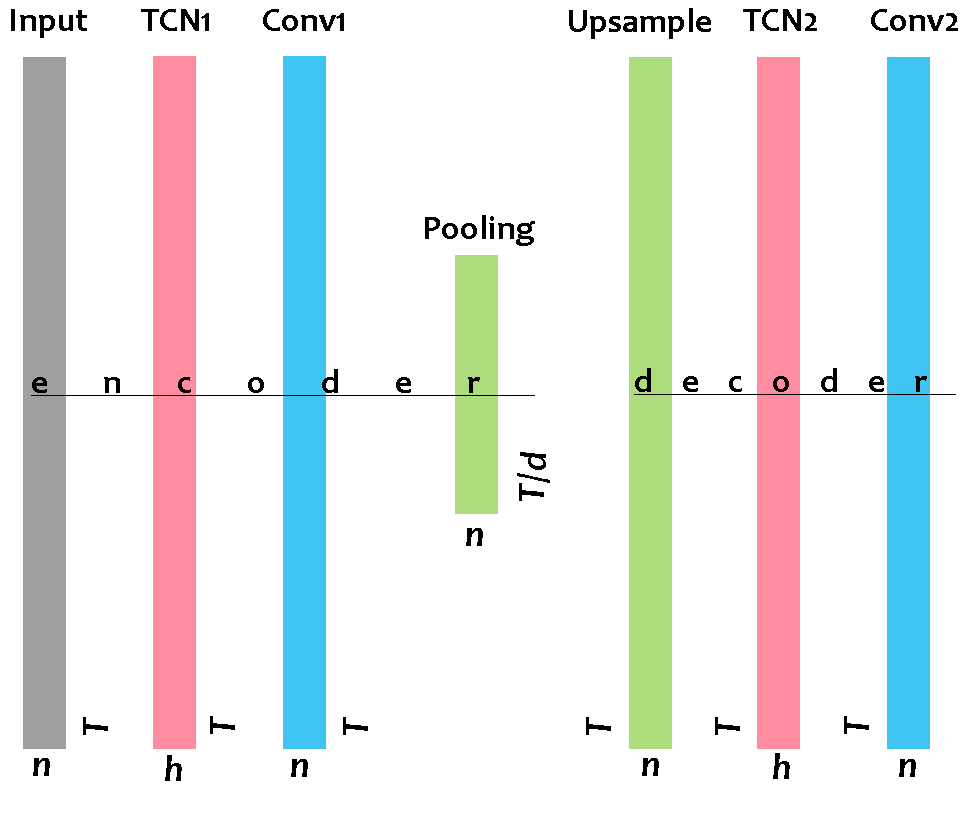}
\caption{The architecture of the TCN-AE for disengagement detection. The input to the TCN-AE is a sequence with length $T$ and dimensionality $n$. TCN1 and TCN2 are TCNs in encoder and decoder parts of TCN-AE with the same architecture but untied/independent weights. Both TCNs have $h$ hidden units and convert the input $T \times n$ sequence to a $T \times h$ sequence. Conv1 and Conv2 are $1 \times 1$ convolutional layers with $h$ input channels and $n$ output channels, thus convert the $T \times h$ sequences to $T \times n$ sequences. The average-pooling bottleneck layer has a kernel of size $d$, outputting a $T/d \times n$ sequence. The upsample layer up-samples the feature maps with a factor of $d$ to restore a $T \times n$ sequence passed through TCN2 and Conv2 to generate the reconstructed version of the input sequence of size $T \times n$.}
\label{fig:tcnae}
\end{figure*}

Thill et al. \cite{thill2021temporal} introduced TCN-AE for unsupervised anomaly detection in time series. As depicted in Figure \ref{fig:tcnae}, a TCN-AE consists of an encoder for compressing (encoding) the input $n$-dimensional feature vector of length $T$ along the time and feature axes. The TCN-AE decoder then attempts to decode the compressed representation and reconstruct the original input feature vector. The architecture of the TCNs in the encoder (TCN1) and decoder (TCN2) are identical and are the vanilla TCN with dilated convolutions \cite{bai2018empirical}. However, the weights of the TCNs are updated independently during training. The Conv1 and Conv2 are trainable $1\times1$ convolution layers. The average-pooling bottleneck layer down-samples the feature map along the time axis. The upsampling layer then up-samples and restores the length of the original feature vector. The up-sampled feature vector is passed through the TCN2 and Conv2 to reconstruct the original feature vector with dimension $n$ and length $T$.

The Mean Squared Error (MSE) is used as the loss function in the training phase. The TCN-AE neural network is trained only by the normal (engaged) samples and is forced to minimize the reconstruction error for these samples. In the test phase, the MSE between the input feature vector and its reconstructed version by the trained TCN-AE is calculated. High values of MSE are expected for the anomalous (disengaged) samples that significantly differ from the normal (engaged) samples on whom the TCN-AE is trained, see Figure \ref{fig:methodology}. As will be explained in Section \ref{sec:experiments}, in addition to the TCN-AE, LSTM- and feedforward-AEs are also examined for disengagement detection in virtual learning.

\section{Experiments}
\label{sec:experiments}
In this section, the performance of the proposed disengagement detection approach is evaluated. There is no previous work on disengagement detection as an anomaly detection problem to be compared with the proposed approach. Therefore, the performance of different AE architectures with different feature sets for disengagement detection is evaluated compared to various binary classification neural networks to show the effectiveness of defining disengagement detection as an anomaly detection problem. A comparison is also made between the proposed method and some of the existing end-to-end and feature-based methods for engagement binary classification.

The evaluation metrics are the Area Under the Curve of the Receiver Operating Characteristic curve (AUC ROC) and the Area Under the Curve of the Precision-Recall curve (AUC PR). The confusion matrix, after finding an optimal threshold on the output reconstruction errors of AEs and on the output class probabilities of binary classifiers, is also calculated (see Section \ref{sec:experimental_result}).

As described in Section \ref{sec:methodology}, while the AEs are trained on only normal (engaged) samples, the binary classifiers are trained on both normal (engaged) and anomalous (disengaged) samples. The test sets of both AE and binary classifiers are the same and contain both normal and anomalous samples.

The experiments were implemented in PyTorch \cite{paszke2019pytorch} and Scikit-learn \cite{pedregosa2011scikit} on a server with 64 GB of RAM and NVIDIA TeslaP100 PCIe 12 GB GPU. The code of our implementations is available at
\url{https://github.com/abedicodes/ENG-AE}.
% \url{https://github.com/xxxxxxx/xxxxxxx}.

\subsection{Dataset}
\label{sec:dataset}
The performance of the proposed method is evaluated on two publicly available video-only student engagement datasets, DAiSEE \cite{gupta2016daisee} and EmotiW \cite{kaur2018prediction}.

\noindent
\textbf{DAiSEE:} The DAiSEE dataset \cite{gupta2016daisee} contains 9,068 videos captured from 112 students in online courses, see Figure \ref{fig:exemplary} for some examples. The videos were annotated by four states of students while watching online courses, boredom, confusion, frustration, and engagement. Each state is in one of the four levels (ordinal classes), level 0 (very low), 1 (low), 2 (high), and 3 (very high). The length, frame rate, and resolution of the videos are 10 seconds, 30 frames per second, and 640 × 480 pixels. In this paper, the focus is only on disengagement detection. Therefore, the videos with engagement levels of 0 and 1 are considered as disengaged (anomalous or positive) samples, and the videos with engagement levels of 2 and 3 are considered as engaged (normal or negative) samples. The number of disengaged and engaged samples in training, validation, and test sets of the DAiSEE dataset is shown in Table \ref{tab:distribution_daisee}. As can be seen in Table \ref{tab:distribution_daisee}, the dataset is highly imbalanced.

\noindent
\textbf{EmotiW:} The EmotiW dataset \cite{kaur2018prediction} contains videos of 78 students in online classroom setting. The total number of videos is 195, including 147 training and 48 validation video samples. The videos are at a resolution of 640 × 480 pixels and 30 fps. The lengths of the videos are around 5 minutes. The engagement levels of students in the videos are in the range 0, 0.33, 0.66, and 1, corresponding to the lowest to highest levels of engagement, where 0, and 1 indicate that the person is completely disengaged, and highly engaged, respectively. Two dichotomizations are applied to use EmotiW for disengagement detection. In the first setting (EmotiW1), the samples with engagement level of 0, and the samples with engagement levels of 0.33, 0.66, and 1.0 are considered as disengaged, and engaged samples, respectively. In the second setting (EmotiW2), the samples with engagement levels of 0 and 0.33, and the samples with engagement levels of 0.66 and 1.0 are considered as disengaged, and engaged samples, respectively, see Table \ref{tab:distribution_emotiw}.

\begin{table}[ht]
\caption{The distribution of engaged and disengaged samples in the train, validation, and test sets in the DAiSEE dataset \cite{gupta2016daisee} after dichotomization of engagement levels in the original dataset, see Section \ref{sec:dataset}.}
\label{tab:distribution_daisee}
\centering
\begin{tabular}{p{.2\linewidth}p{.15\linewidth}p{.15\linewidth}p{.15\linewidth}}
\hline
Engagement&Train&Validation&Test\\
\hline
disengaged&247&166&88
\\
\hline
engaged&5111&1263&1696
\\
\hline
total&5358&1429&1784
\\
\hline
\end{tabular}
\end{table}

\begin{table}[ht]
\caption{The distribution of engaged and disengaged samples in the train and validation sets in the EmotiW dataset \cite{kaur2018prediction} after dichotomization of engagement levels in the original dataset in two different ways resulting in EmotiW1 and EmotiW2, see Section \ref{sec:dataset}.}
\label{tab:distribution_emotiw}

\begin{tabular}{p{.15\linewidth}p{.1\linewidth}p{.1\linewidth}p{.1\linewidth}p{.1\linewidth}p{.15\linewidth}p{.1\linewidth}p{.1\linewidth}}

\hline
& \multicolumn{2}{c}{EmotiW1} & \multicolumn{3}{c}{EmotiW2}\\
\hline
Engagement&Train&Validation & & Train&Validation\\
\hline
disengaged&5&4&&40&14
\\
\hline
engaged&142&44&&107&34
\\
\hline
total&147&48&&147&48
\\
\hline
\end{tabular}

\end{table}

\subsection{Experimental Setting}
\label{sec:experimental_setting}
The frame-level behavioral features, described in Section \ref{sec:feature_extraction}, are extracted by the OpenFace \cite{baltrusaitis2018openface}. The OpenFace also outputs the extracted face regions from video frames. The extracted face regions of size 256 × 256 are fed to the pretrained EmoFAN \cite{toisoul2021estimation} on AffectNet \cite{mollahosseini2017affectnet} for extracting affect features (see Section \ref{sec:feature_extraction}). The frame-level features are extracted from 10-second video samples (with 300 frames) of the DAiSEE dataset to be fed to the AEs. The extracted feature vector from each frame is considered as one time step of the temporal AE models. As the EmotiW dataset contains videos of around 5-minute length, following the previous works on this dataset \cite{thomas2018predicting,liao2021deep, abedi2021affect}, the videos are divided into 10-second segments with $50\%$ overlap, and segment-level features, described in Section \ref{sec:feature_extraction}, are extracted from each segment. The segment-level feature vector extracted from each video segment is considered as one time step of the temporal AE models.

For the TCN-AE (described in Section \ref{sec:anomaly_detection}), the parameters of the TCN1 and TCN2, giving the best results, are as follows, 8, $h=24$, 8, and 0.05 for the number of levels, number of hidden units, kernel size, and dropout \cite{bai2018empirical}. The Conv1 and Conv2 are $1\times1$ convolutional layers with $h=24$ input channels and $n$ output channels. The average-pooling bottleneck layer has a kernel of size $d=4$. The interpolation layer up-samples the feature maps with a factor of $d=4$. $n$ is the dimensionality or the number of features in the feature vector. In frame-level features, the number of features \textit{n} is 9, and 11 when using only behavioral, and both behavioral and affect features, respectively. In segment-level features, the number of features \textit{n} is 33, and 37 when using only behavioral, and both behavioral and affect features, respectively. To compare the perfomance of the TCN-AE with the TCN binary classifier, one TCN with the same architecture as TCN1, trailed by a $24\times1$ fully-connected layer and a Sigmoid activation function is also implemented. The binary cross-entropy loss is used as the loss function for the TCN binary classifier.

In addition to the TCN-AE, LSTM- and feedforward AEs are also implemented as follows. The encoder part of the LSTM AE contains two LSTMs. The first LSTM accepts the \textit{n}-element feature vector and the number of neurons in its hidden layer is \textit{h}. The second LSTM’s input dimension is \textit{h} and the number of neurons in its hidden layer is \textit{b}. \textit{b} is the dimension of the encoded version of input, i.e., the output of encoder. The decoder part of the LSTM AE contains two LSTMs. The first LSTM’s input, and output dimensions are \textit{b}, and \textit{h}, while the second LSTM’s input, and output dimensions are \textit{h}, and \textit{n}, respectively. The \textit{n}-dimensional output of the final LSTM is the reconstruction of the input. To compare the performance of AEs with a binary classifier, the encoder part of the LSTM AE trailed by a fully connected layer and a Sigmoid has also been implemented. The binary cross-entropy loss is used as the loss function for the binary classifier. The feedforward AE contains two fully-connected layers as encoder and two fully-connected layers as decoder of size, \textit{n} × (2 × \textit{b}), (2 × \textit{b}) × \textit{b}, \textit{b} × (2 × \textit{b}), and (2 × \textit{b}) × \textit{n}. The values of \textit{b}, and \textit{h} are 64, and 128 achieving the best results.

To compare the performance of the above feature-based AEs with end-to-end models, a 3D-Convolutional Neural Network (CNN) AE is also implemented. The encoder part of the 3D-CNN AE contains three 3D convolution, ReLU, and 3D max-pooling blocks. The decoder part contains three 3D transposed convolution, ReLU, and 3D max-pooling blocks to reconstruct the input. For comparison, several end-to-end binary classifiers are also implemented. See the references in Table \ref{tab:literature} for more details. In all the models, the Adam optimizer with decaying learning rate is used for optimization.

The majority of the existing engagement measurement approaches solve multi-class classification or regression problems, according to Table \ref{tab:literature}. To compare the binary classification version of those methods with the proposed method, the final fully-connected layer of their proposed neural network architecture is altered to have one output neuron (for binary classification), and their loss function is changed to binary cross-entropy loss.

\subsection{Experimental Result}
\label{sec:experimental_result}
Table \ref{tab:results_daisee} shows the AUC ROC and AUC PR of different models for disengagement detection of the test samples in the DAiSEE dataset. Table \ref{tab:results_emotiw} shows the AUC ROC and AUC PR of different models for disengagement detection of the validation samples in the EmotiW1 and EmotiW2 settings of the EmotiW dataset. As the main focus of this paper is studying the performance of AEs in disengagement detection as an anomaly detection problem, for each AE model, the result of its binary classification counterpart is also reported. The results of using two feature vectors of behavioral features only and behavioral and affect features together are reported for each model. The higher AUC ROC and AUC PR of the AE models, including feedforward-, LSTM-, and TCN-AEs, compared to their equivalent binary classifiers (described in Section \ref{sec:experimental_setting}) show the effectiveness of detecting disengagement in an anomaly detection setting. In almost all the models, after adding the affect features to the behavioral features, the performance improves. It shows the effectiveness of the affect features as indicators of engagement (or disengagement). The superiority of the temporal LSTM- and TCN- models, analyzing sequences of feature vectors, over non-temporal feedforward models, shows the importance of analyzing the temporal changes in the affect and behavioral states for disengagement detection. In all the DAiSEE, EmotiW1, and EmotiW2, TCN-AE significantly outperforms LSTM-AE. This is because of the superiority of TCN in modeling sequences of larger length and retaining memory of history compared to the LSTM.

\begin{table*}[ht]
\caption{AUC ROC and AUC PR of different sets of features and different neural network models for disengagement detection on the test set of the DAiSEE dataset, see Section \ref{sec:experimental_result}. BC, and AE are Binary Classifier, and Auto Encoder, respectively.}
\label{tab:results_daisee}
\centering
\begin{tabular}{p{.25\linewidth}p{.35\linewidth}p{.1\linewidth}p{.1\linewidth}}
\hline
Feature & Model & AUC ROC & AUC PR\\
\hline
end-to-end & C3D + LSTM-BC \cite{abedi2021improving} & 0.5931 & 0.1058 \\ \hline
end-to-end & C3D + TCN-BC \cite{abedi2021improving} & 0.6076 & 0.1277 \\ \hline
end-to-end & ResNet + LSTM-BC \cite{abedi2021improving} & 0.6243 & 0.146 \\ \hline
end-to-end & ResNet + TCN-BC \cite{abedi2021improving} & 0.6084	& 0.1409\\ \hline
end-to-end & 3D CNN-BC \cite{gupta2016daisee} & 0.6022 & 0.1131\\ \hline
end-to-end & 3D CNN-AE & 0.6119 & 0.1501\\ \hline

behavioral & feedforward-BC \cite{abedi2021affect} & 0.6855 & 0.1262\\ \hline
behavioral + affect & feedforward-BC \cite{abedi2021affect} & 0.6837 & 0.1141\\ \hline

behavioral & feedforward-AE & 0.7093 & 0.1594\\ \hline
behavioral + affect & feedforward-AE & 0.7171 & 0.2023\\ \hline

behavioral & LSTM-BC \cite{abedi2021affect} & 0.6742 & 0.1675\\ \hline
behavioral + affect & LSTM-BC \cite{abedi2021affect} & 0.6918 & 0.1556\\ \hline

behavioral & LSTM-AE & 0.7007 & 0.1972\\ \hline
behavioral + affect & LSTM-AE & 0.7733 & 0.2400\\ \hline

behavioral & TCN-BC \cite{abedi2021affect} & 0.6755 & 0.2027\\ \hline
behavioral + affect & TCN-BC \cite{abedi2021affect} & 0.7160 & 0.2182\\ \hline
behavioral + affect & TCN-BC (WL) \cite{abedi2021affect} & 0.7516 & 0.2007\\ \hline

behavioral & TCN-AE & 0.7600 & 0.2324\\ \hline
behavioral + affect & TCN-AE & \textbf{0.7974} & \textbf{0.2632}\\ \hline

\hline
\end{tabular}
\end{table*}

\begin{table*}[ht]
\caption{AUC ROC and AUC PR of different sets of features and different neural network models for disengagement detection on the validation set of the EmotiW dataset in EmotiW1 and EmotiW2 settings, see Section \ref{sec:experimental_result}. BC, and AE are Binary Classifier, and Auto Encoder, respectively.}
\label{tab:results_emotiw}
\centering
\begin{tabular}{p{.15\linewidth}p{.25\linewidth}p{.25\linewidth}p{.075\linewidth}p{.075\linewidth}}
\hline
Setting & Feature & Model & AUC ROC & AUC PR\\
\hline

EmotiW1 & behavioral & LSTM-BC \cite{abedi2021affect} & 0.7811 & 0.3111\\ \hline
EmotiW1 & behavioral + affect & LSTM-BC \cite{abedi2021affect} & 0.8077 & 0.3302\\ \hline

EmotiW1 & behavioral & LSTM-AE & 0.9012 & 0.3494\\ \hline
EmotiW1 & behavioral + affect & LSTM-AE & 0.907 & \textbf{0.5028}\\ \hline

EmotiW1 & behavioral & TCN-BC \cite{abedi2021affect} & 0.8372 &0.301\\ \hline
EmotiW1 & behavioral + affect & TCN-BC \cite{abedi2021affect} & 0.8488 & 0.3078\\ \hline
EmotiW1 & behavioral + affect & TCN-BC (WL) \cite{abedi2021affect} & 0.8641 & 0.3127\\ \hline

EmotiW1 & behavioral & TCN-AE & 0.9128 & 0.3312\\ \hline
EmotiW1 & behavioral + affect & TCN-AE & \textbf{0.936} & 0.4125\\ \hline
\hline

EmotiW2 & behavioral & LSTM-BC \cite{abedi2021affect} & 0.6883 & 0.3908\\ \hline
EmotiW2 & behavioral + affect & LSTM-BC \cite{abedi2021affect} & 0.7056 & 0.4035\\ \hline

EmotiW2 & behavioral & LSTM-AE & 0.7424 & 0.4451\\ \hline
EmotiW2 & behavioral + affect & LSTM-AE & 0.7446 & 0.4448\\ \hline

EmotiW2 & behavioral & TCN-BC \cite{abedi2021affect} & 0.7121 & 0.4148\\ \hline
EmotiW2 & behavioral + affect & TCN-BC \cite{abedi2021affect} & 0.7013 &  0.4077\\ \hline
EmotiW2 & behavioral + affect & TCN-BC (WL) \cite{abedi2021affect} & 0.7169 &  0.4215\\ \hline

EmotiW2 & behavioral & TCN-AE & 0.7403 & 0.4445\\ \hline
EmotiW2 & behavioral + affect & TCN-AE & \textbf{0.7489} & \textbf{0.4582}\\ \hline

\hline
\end{tabular}
\end{table*}

For the DAiSEE in Table \ref{tab:results_daisee}, the results of end-to-end binary classification models are also reported for comparison (see \cite{abedi2021improving} for details of the end-to-end models). Due to the longer length of videos, these models are not applicable to EmotiW. While the performance of 3D CNN AE is better than 3D CNN binary classifier and other end-to-end models, the performance of the end-to-end models is much lower than feature-based models. Therefore, the behavioral and affect handcrafted features are more successful in differentiation between engagement and disengagement.

The baseline value for AUC PR (the performance of a random binary classifier) is the ratio of the number of positive samples to the total number of samples \cite{khan2022unsupervised}. Referring to Table \ref{tab:distribution_daisee} and Table \ref{tab:distribution_emotiw}, this baseline value is 88 / 1784 = 0.0493, 4 / 48 = 0.0833, and 14 / 48 =  0.2917 for DAiSEE, EmotiW1, and EmotiW2, respectively. As can be seen in Table \ref{tab:results_daisee} and \ref{tab:results_emotiw}, all the developed models outperform the random binary classifier by a large margin. Also, the AUC ROC of the developed models, specially AEs, is much higher than the AUC ROC of a random classifier (0.5).

Due to the imbalanced distribution of samples in the datasets \ref{sec:dataset}, we reimplement the best binary classifiers (using both behavioral and affect features and TCN) using a weighted binary cross-entropy loss function \cite{abedi2021improving,paszke2019pytorch}. According to Tables 4 and 5, the binary classifiers with weighted loss functions (indicated by \textit{WL}) outperform their non-weighted counterparts; however, their performance remains inferior to that of AEs.

Comparing the performance of the TCN-AE and TCN binary classifier using behavioral and affect features for the DAiSEE, EmotiW1, and EmotiW2 datasets in Table \ref{tab:results_daisee} and Table \ref{tab:results_emotiw}, the percentages of improvement in AUC ROC are 11.37\%, 10.27\%, and 6.78\%, respectively, and the percentages of improvement in AUC PR are 20.62\%, 34.01\%, and 12.39\%, respectively. Therefore, the average improvement of the TCN-AE, compared to the TCN binary classifier, is $9$\% on AUC ROC and $22$\% on AUC PR.

The confusion matrices in Table \ref{tab:confusion_matrices} are the results of thresholding reconstruction errors and output class probabilities of TCN-AE and TCN binary classifier using behavioral and affect features on the DAiSEE, EmotiW1, and EmotiW2 datasets. As the TCN-AE is trained in an unsupervised setting and only on the engaged (normal) samples, the disengaged (anomalous) samples cannot be used for finding an optimal threshold for thresholding reconstruction errors \cite{thill2021temporal,khan2017detecting}.
% Also, the EmotiW dataset does not contain a separate set for validation to find an optimal threshold.
Therefore, the instructions in \cite{khan2017detecting} are followed to find an optimal threshold using only engaged (normal) samples of the training set. The numbers in the confusion matrices are the number of correctly and incorrectly classified samples. In all the DAiSEE, EmotiW1, and EmotiW2 datasets, TCN-AE outperforms TCN binary classifier in detecting disengaged samples.

\begin{table*}[ht]
\caption{Confusion matrices of the TCN-AE and TCN binary classifier (TCN-BC) using behavioral and affect features on DAiSEE, EmotiW1, and EmotiW2 after thresholding reconstruction error of TCN-AE and output class probabilities of TCN-BC. Eng., and Dis. are Engaged (negative), and Disengaged (positive), respectively. The numbers are the number of correctly and incorrectly classified samples, (starting from Eng. and Eng. and clockwise) corresponding to true-negative, false-positive, true-positive, and false-negative. see Section \ref{sec:experimental_result}.}
\label{tab:confusion_matrices}
\centering
\begin{tabular}{
p{.15\linewidth}
p{.20\linewidth}
p{.25\linewidth}
p{.25\linewidth}}
\hline
 & DAiSEE
 & EmotiW1 & EmotiW2\\
\hline
TCN-BC
&
\begin{tabular}{p{.1\linewidth}p{.2\linewidth}p{.2\linewidth}}
\hline
         & Eng. & Dis.\\
        \hline
        Eng. & 1254 & 442
        \\
        \hline
        Dis. & 51 & 37
        \\
        \hline
\end{tabular}
&
\begin{tabular}{p{.1\linewidth}p{.2\linewidth}p{.2\linewidth}}
\hline
         & Eng. & Dis.\\
        \hline
        Eng. & \textbf{35} & 9
        \\
        \hline
        Dis. & 1 & 3
        \\
        \hline
\end{tabular}
&
\begin{tabular}{p{.1\linewidth}p{.2\linewidth}p{.2\linewidth}}
\hline
         & Eng. & Dis.\\
        \hline
        Eng. & 26 & 8
        \\
        \hline
        Dis. & 9 & 5
        \\
        \hline
\end{tabular}

\\ \hline

TCN-AE
&
\begin{tabular}{p{.1\linewidth}p{.2\linewidth}p{.2\linewidth}}
\hline
         & Eng. & Dis.\\
        \hline
        Eng. & \textbf{1273} & 423
        \\
        \hline
        Dis. & 38 & \textbf{50}
        \\
        \hline
\end{tabular}
&
\begin{tabular}{p{.1\linewidth}p{.2\linewidth}p{.2\linewidth}}
\hline
         & Eng. & Dis.\\
        \hline
        Eng. & 32 & 12
        \\
        \hline
        Dis. & 0 & \textbf{4}
        \\
        \hline
\end{tabular}
&
\begin{tabular}{p{.1\linewidth}p{.2\linewidth}p{.2\linewidth}}
\hline
         & Eng. & Dis.\\
        \hline
        Eng. & \textbf{27} & 7
        \\
        \hline
        Dis. & 3 & \textbf{11}
        \\
        \hline
\end{tabular}

\\ \hline
\end{tabular}
\end{table*}

\section{Conclusion and Future Work}
\label{sec:conclusion}
In this paper, for the first time, we defined and addressed the problem of detecting student disengagement in videos as an anomaly detection problem. We designed various AE neural networks, including TCN-AE \cite{thill2021temporal}, using different behavioral and affect features and evaluated their performance on two publicly available student engagement measurement datasets, DAiSEE \cite{gupta2016daisee} and EmotiW \cite{kaur2018prediction}. The proposed approach successfully detects disengagement with high AUC ROC and AUC PR values compared to binary engagement/disengagement classifiers. In future work, we plan to investigate incorporating attention mechanism into the architecture of AEs for disengagement detection to make them pay attention to significant behavioral and affect states and temporal changes in the behavioral and affect states that are indicators of disengagement and to make the developed AEs more interpretable. We plan to explore supervised and unsupervised contrastive learning approaches \cite{kopuklu2021driver,khan2021modified} for video-based engagement/disengagement measurement. We aim to collect video datasets in virtual rehabilitation settings and analyze the effectiveness of the proposed approach to measure engagement in other types of virtual learning environments.
\\\\
\textbf{Data availability}\\
The datasets analyzed during the current study are publicly available in the following repositories:\\
https://people.iith.ac.in/vineethnb/resources/daisee/index.html\\
https://sites.google.com/view/emotiw2020/
\\\\
\textbf{Conflict of Interest}\\
The authors declare that they have no conflict of interest.

\bibliography{sn-bibliography}

\end{document}